\documentclass{article} 

\usepackage[preprint]{neurips_2026}

\usepackage{amsmath,amsfonts,bm}

\def\eqref#1{equation~\ref{#1}}

\def\1{\bm{1}}

\DeclareMathAlphabet{\mathsfit}{\encodingdefault}{\sfdefault}{m}{sl}
\SetMathAlphabet{\mathsfit}{bold}{\encodingdefault}{\sfdefault}{bx}{n}

\usepackage[hidelinks]{hyperref}
\usepackage{url}
\usepackage{booktabs}
\usepackage{graphicx}
\usepackage{wrapfig}
\usepackage{multirow}
\usepackage{xspace}

\newcommand{\score}{
  \ifmmode\mathrm{PARA}
  \else\textsc{PARA}\xspace
  \fi
}

\newcommand{\model}{
  \ifmmode\mathrm{GRAD}
  \else\textsc{GRAD}\xspace
  \fi
}

\title{
Perception or Reasoning? Diagnosing and Tackling
Reinforcement Learning Bottlenecks in VLMs
}

\author{
\makebox[0.98\textwidth][c]{
\textbf{Zeyuan Yang$^{1,\dagger}$}
\hspace{0.1em}
\textbf{Zhuohao Chen$^{1,\dagger}$}
\hspace{0.1em}
\textbf{Jiaben Chen$^{1}$}
\hspace{0.1em}
\textbf{Yao Zhang$^{2}$}
\hspace{0.1em}
\textbf{Maohao Shen$^{3}$}
\hspace{0.1em}
\textbf{Chuang Gan$^{1}$}
}\\[0.65em]
\makebox[0.98\textwidth][c]{
{\large
$^{1}$ UMass Amherst
\hspace{1.5em}
$^{2}$ UIUC
\hspace{1.5em}
$^{3}$ MIT}
}
}

\makeatletter
\renewcommand{\@notice}{}
\makeatother

\newcommand{\authorfootnote}[1]{
  \begingroup
  \renewcommand{\thefootnote}{}
  \footnote{#1}
  \addtocounter{footnote}{-1}
  \endgroup
}

\begin{document}

\maketitle
\authorfootnote{
$^\dagger$ Equal contribution.
}

\begin{abstract}
Reinforcement learning has brought strong gains to language model reasoning, but its improvements on visual tasks remain less consistent. One difficulty is that visual task failures can arise from different sources: the same benchmark may be limited by perception for one model and by reasoning for another. We introduce Perception and Reasoning Attribution (PARA), which measures the remaining perceptual and reasoning headroom of each model--benchmark pair by comparing how many current errors are recovered with more accurate visual information versus a stronger reasoner. Across 4 vision-language models and 17 multimodal benchmarks, PARA broadly agrees with the expected characteristics of existing benchmarks while revealing substantial model-specific variation. More importantly, it shows that RL gains become smaller as perceptual headroom increases, but larger when reasoning explains more of the recoverable errors, suggesting that current RL exploits reasoning headroom more effectively than perceptual headroom. Motivated by this gap, we analyze how existing RL objectives affect visual perception and find that reward- and token-level signals influence visual attention only indirectly. We therefore propose GRAD, a group-relative attention distillation objective that uses the best rollout in each group to directly guide the visual attention of weaker rollouts. GRAD consistently improves performance across two model scales and seven benchmarks, and further complements existing RL methods.
\end{abstract}
\section{Introduction}

\begin{figure}[t]
    \centering
    \includegraphics[width=0.9\linewidth]{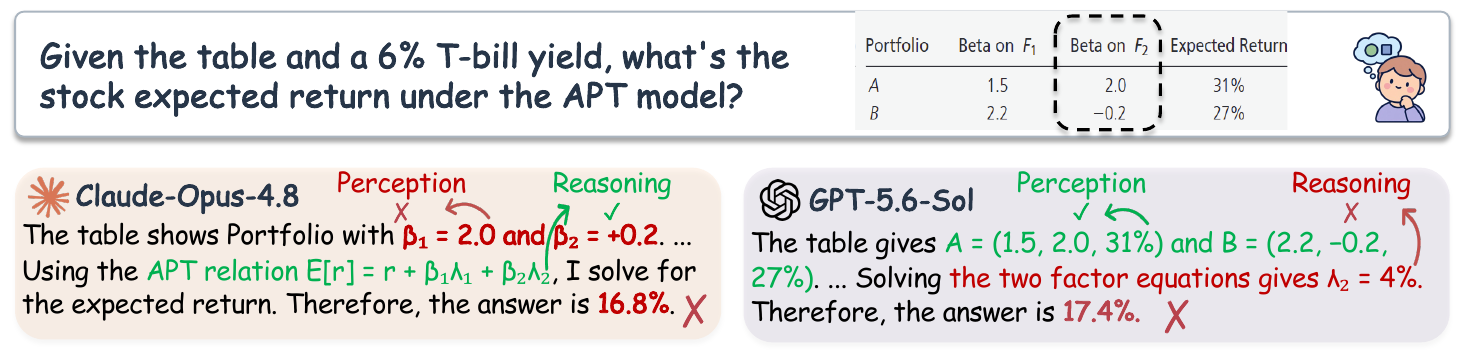}
    \caption{\textbf{The same task can expose different failure modes across models.}
    One model may fail to perceive the relevant evidence, while another perceives it correctly but fails to reason from it.}
    \label{fig:teaser}
\end{figure}

Reinforcement learning (RL) has become a key post-training paradigm for large language models (LLMs) and vision-language models (VLMs), producing substantial gains in mathematical, coding, and multimodal reasoning tasks~\citep{shao2024deepseekmath,deepseekai2025r1,huang2025visionr1,shen2025vlmr1,liu2025visualrft,yu2025dapo}.
Yet these gains remain less consistent on visual tasks.
A prevailing explanation is that RL improves reasoning more readily than perception, leaving access to visual evidence as a persistent bottleneck~\citep{xiao2025perceptionr1,li2026frankenstein,wu2026asymmetric}.
However, this observation does not reveal what limits a particular model on a particular task.
As illustrated in Fig.~\ref{fig:teaser}, the same problem may fail because a model misperceives the necessary evidence or because it fails to reason from correctly perceived evidence.
Existing analyses largely rely on benchmark characteristics, controlled interventions, or manual inspection~\citep{wang2025reasoningbenchmark,li2026frankenstein,wu2026asymmetric}, leaving no general way to quantify how much improvement remains available from perception versus reasoning at the model--benchmark level.

Motivated by this gap, we introduce \emph{Perception and Reasoning Attribution} (\score), an intervention-based framework that measures how much of a model's current errors is recoverable through stronger perception versus stronger reasoning.
It estimates these two headroom by selectively strengthening visual information or reasoning while preserving the other one.
Across 4 VLMs and 17 benchmarks, \score reveals broad benchmark-level trends alongside substantial model-specific variation.

Leveraging \score, we further reveal a consistent asymmetry in visual RL: gains decrease with perceptual headroom but increase with reasoning-related headroom, suggesting that current RL exploits reasoning more effectively than perception.
We show that reward- and token-level perceptual signals affect visual attention only indirectly, whereas direct attention supervision responds more systematically to perceptual headroom.
Together, these results identify visual attention as a more direct optimization interface for perception.

Direct attention supervision, however, still requires a reliable target because outcome rewards conflate perception and reasoning.
We therefore propose \model, \emph{Group-Relative Attention Distillation}, which uses the best sibling within each on-policy rollout group as a positive visual-attention target for weaker rollouts.
Across two model scales and seven benchmarks, \model consistently improves performance, strengthens perception, and remains complementary to existing RL methods, without requiring external teachers or additional annotations.
Our contributions are threefold:
\begin{itemize}
    \item \textbf{Diagnosing model-specific perception and reasoning bottlenecks.}
    We introduce \score, an intervention-based measure of recoverable perceptual and reasoning headroom.
    Across four VLMs and 17 benchmarks, \score reveals substantial model-specific variation that cannot be inferred from benchmark labels alone.

    \item \textbf{Explaining perception--reasoning asymmetry of visual RL.}
    We show that RL gains decrease with perceptual headroom but increase with reasoning, and connect this asymmetry to indirect optimization path from reward- or token-level signals to visual attention.

    \item \textbf{Tackling perceptual bottlenecks with group-relative attention distillation.}
    We propose \model, which distills visual attention from the best on-policy sibling into weaker rollouts.
    \model improves performance across models and benchmarks without external teachers or additional annotations, while remaining complementary to existing RL objectives.
\end{itemize}
\section{Related Work}

\paragraph{Post-Training Vision-Language Models.} Post-training has become central to adapting vision-language models for instruction following and complex reasoning. Recent work extends reinforcement learning with verifiable rewards from language models to VLMs, combining GRPO-style optimization with answer- and format-based rewards, task-specific verifiers, multimodal cold starts, or structured visual reasoning traces~\citep{shao2024deepseekmath, deepseekai2025r1,yu2025dapo,huang2025visionr1,shen2025vlmr1,liu2025visualrft,xia2025visionaryr1,yang2025wethink,lu2026vico,zeng2026scalable}. Although these approaches substantially improve downstream performance, their outcome-level feedback does not identify whether a failure originates from perception or reasoning. Recent methods therefore introduce more informative perception supervision, including visual-description rewards, perception checklists, explicit visual--textual grounding signals, and perception-centric process rewards ~\citep{xiao2025perceptionr1,li2025visionsr1,ding2025vtperception,zhang2026pearl,min2026perceval,
kumar2026splitrl,li2026umrm,tian2026vrubrics}, or perception-first training curricula~\citep{wu2026seeing}. Other methods refine the learning signal through visual-dependency-aware trajectory and token weighting~\citep{huang2025spotlight,wang2026vgpo,li2026prpo,wang2026ttrsd}, or regularize output distributions under altered visual evidence~\citep{wang2026papo,zhang2026bips,rezaei2026cppo}. Controlled studies likewise show that post-training improves reasoning more readily than perception, attributing this asymmetry in RL to the weak coupling between rewards and perception~\citep{wu2026asymmetric,
ye2026samereward,zhou2026robustmmrl}. These reward- and token-level objectives act through scalar returns, policy gradients, or output distributions and are complementary to our attention-level objective. Our work measures the perception and reasoning bottlenecks of each model--benchmark pair and uses this diagnosis to directly optimize visual-attention distributions. Complementary work intervenes in vision representations or supervises visual-token attention~\citep{song2026rlsee, li2026frankenstein,song2026gaze,li2026ral,hu2026ease,berger2026selfgrounded,song2026anchoring}. Unlike formulations that rely on a separate teacher or annotated visual evidence, our attention targets are constructed from the model's own on-policy rollout groups.

\paragraph{On-Policy Distillation.} Conventional knowledge distillation matches a student's output distributions or sequences to those of a teacher on fixed data~\citep{hinton2015distilling,kim2016sequence}. MiniLLM and GKD instead distill teacher logits on student-generated trajectories, reducing the discrepancy between training and inference distributions~\citep{gu2024minillm,agarwal2024gkd}. This logit-level formulation has been extended to VLMs: VOLD transfers reasoning from a text-only teacher~\citep{bousselham2026vold}; Vision-OPD uses a crop-conditioned model as a privileged teacher~\citep{yuan2026visionopd}; and VA-OPD and ViCuR focus distillation on visually dependent tokens or recoverable visual cues~\citep{liu2026vaopd,tian2026vicur,zhang2026redraft}. Attention provides a complementary distillation target because it explicitly captures how output tokens attend to visual evidence. OPAD jointly matches a teacher's output logits and internal attention distributions on student rollouts~\citep{li2026ral}, while related attention-supervision methods optimize visual-token routing or align attention with annotated evidence regions~\citep{song2026gaze,hu2026ease}. Most prior formulations therefore require either an expert teacher or additional annotated or privileged data, such as reasoning traces, evidence regions, crops, or visual cues. In contrast, our method performs group-relative attention distillation using targets constructed within each rollout group, without a separate teacher or extra region and rationale annotations.
\section{Diagnosing Perception and Reasoning Bottlenecks in Vision-Language Model RL}

Visual task failures can arise from insufficient perception, insufficient reasoning, or both.
Importantly, the dominant limitation is not determined by the benchmark alone: the same task may expose different bottlenecks for different models.
We therefore introduce \score, a model--benchmark specific framework for measuring how much of a model's current error set is recoverable through stronger perception versus stronger reasoning.
We first define these two forms of recoverable headroom, then validate the description-based proxy underlying the measurement, and finally use \score to characterize how perception and reasoning bottlenecks vary across models and benchmarks.

\subsection{\score: Perception and Reasoning Attribution}
\label{subsec:para}
\begin{figure}[t]
\centering
\includegraphics[width=\linewidth]{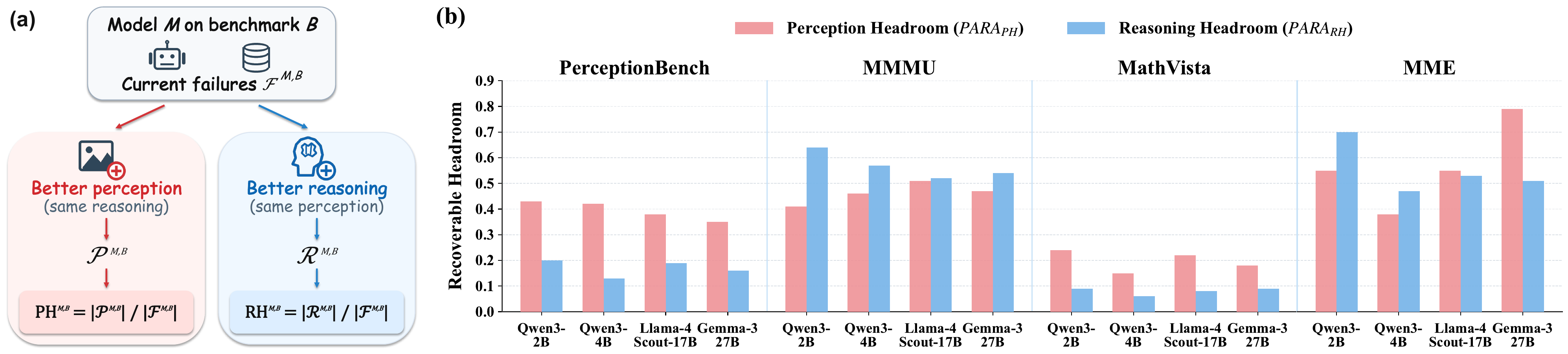}
\caption{\textbf{(a) Illustration of \score.} Starting from a model's current failures, \score measures the fractions recoverable by selectively strengthening perception or reasoning.
\textbf{(b) Representative Headroom Measurements.} Results exhibit broad benchmark-level trends but substantial model-specific variation, showing that effective bottlenecks cannot be inferred by benchmark labels alone.}
\label{fig:para_main}
\end{figure}

Visual Benchmarks are often characterized by the capabilities they are designed to assess, but such labels do not identify what limits a particular model.
For example, one model may fail on a visual reasoning problem because it misperceives a relevant object, whereas another may perceive the same evidence correctly but fail to reason from it.
We therefore define \score at the model--benchmark level and measure two intervention-based quantities: \emph{perceptual headroom}, corresponding to failures recoverable with stronger visual information while retaining the target model's own reasoning, and \emph{reasoning headroom}, corresponding to failures recoverable with stronger reasoning while retaining the target model's own perceived visual information.
Fig.~\ref{fig:para_main}(a) illustrates this construction.

Formally, let $F^{M,B}$ denote the set of examples in benchmark $B$ that model $M$ answers incorrectly, and let $n_F^{M,B} = |F^{M,B}|$.
We estimate the two forms of headroom by selectively strengthening one capability while preserving the other as much as possible.

\textbf{Perceptual Intervention.}
To estimate perceptual headroom, a stronger reference model $M^{o,i}$ generates a detailed description of the image without seeing the question.
The target model $M$ then answers the original question using this additional visual information.
The subset of current failures corrected under this intervention is denoted
\begin{equation}
    P_i^{M,B} \subseteq F^{M,B}.
\end{equation}
To reduce sensitivity to information loss or noise in the description, we evaluate the target model both from the description alone and from the original image augmented with the same description, and count a failure as recovered if either condition succeeds.

\textbf{Reasoning Intervention.}
To estimate reasoning headroom, we reverse the intervention.
The target model $M$ first generates its own question-blind description of the image, which serves as a proxy for the visual information it has retained.
The stronger reference model $M^{o,i}$ then answers the original question using this description.
The corresponding recovered failures form
\begin{equation}
    R_i^{M,B} \subseteq F^{M,B}.
\end{equation}
We include an example in $R_i^{M,B}$ only when the reference model can also answer it correctly from the original image.
This solvability check prevents failures beyond the reference model's own capability from being incorrectly attributed to the target model's reasoning.
Using multiple reference models, we aggregate the recovered examples as
\begin{equation}
    P^{M,B}
    =
    \bigcup_{i=1}^{|\mathcal{M}^{o}|}
    P_i^{M,B},
    \qquad
    R^{M,B}
    =
    \bigcup_{i=1}^{|\mathcal{M}^{o}|}
    R_i^{M,B}.
    \label{eq:para-recovery}
\end{equation}
We use the term \emph{reference model} operationally: these models provide stronger intervention capabilities but are not assumed to be infallible.
In our experiments, we instantiate
$\mathcal{M}^{o}$
with two frontier APIs,
\texttt{Claude-Opus-4.8}

and
\texttt{GPT-5.6-Luna}

The framework naturally extends to additional or stronger reference models.
We then define the perceptual and reasoning headroom as
\begin{equation}
    \score_{\mathrm{PH}}^{M,B}
    =
    \frac{|P^{M,B}|}{n_F^{M,B}},
    \qquad
    \score_{\mathrm{RH}}^{M,B}
    =
    \frac{|R^{M,B}|}{n_F^{M,B}}.
    \label{eq:para-headroom}
\end{equation}
Where $\score_{\mathrm{PH}}^{M,B}$ measures the fraction of the target model's current failures recoverable through stronger perceptual information, and $\score_{\mathrm{RH}}^{M,B}$ measures the fraction recoverable through stronger reasoning given the visual information retained by the target model.
These quantities measure \emph{recoverable headroom}, rather than assigning an intrinsic perception or reasoning label to the benchmark.

\subsection{Validating Question-Blind Descriptions as a Perception Proxy.}

The reasoning intervention relies on the target model's question-blind description as a proxy for the visual information it has extracted from the image.
Because textual descriptions may discard visual information and thereby impair downstream reasoning~\citep{wang2023imagegap}, we evaluate how faithfully this representation preserves the target model's effective perceptual capability. Specifically, for each model and benchmark, we compare two conditions using the same question-answering instruction.
In the \emph{direct-image} condition, the model answers directly from the original image.
In the \emph{self-description} condition, the model first generates a question-blind description and subsequently answers the question from that description alone.
If the description preserves most of visual information available to the model, performance in two conditions should closely correlate one another.

\begin{figure*}[t]
\centering
\includegraphics[width=\textwidth]{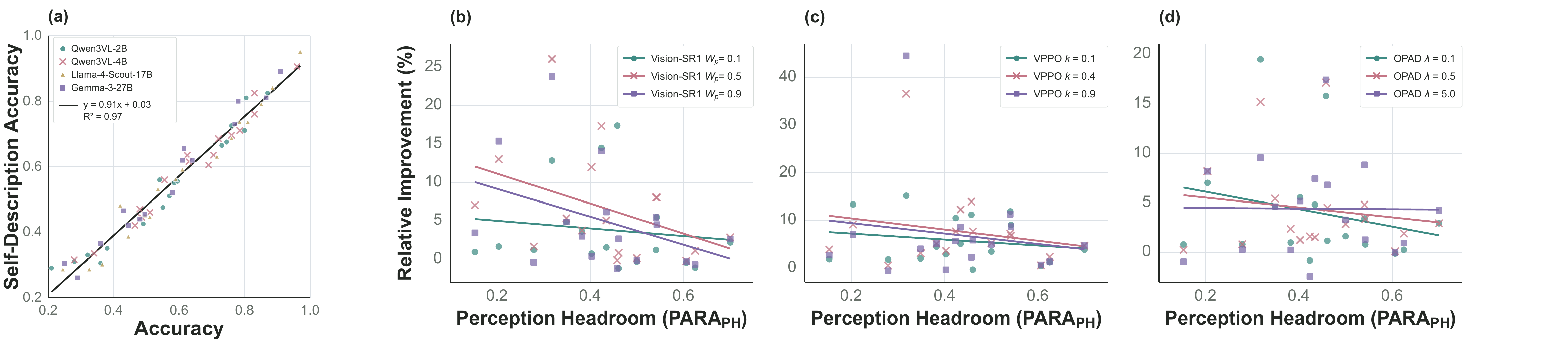}
\caption{\textbf{(a) \score validation.} Self-description accuracy closely tracks direct image accuracy. \textbf{(b)(c)(d) Perceptual supervision comparison.} Only direct attention supervision exhibits a coefficient-dependent response to perceptual headroom $\score_{\mathrm{PH}}$.}
\label{fig:supervision-sweep}
\end{figure*}

As shown in Fig.~\ref{fig:supervision-sweep}(a), self-description accuracy closely follows direct-image accuracy across models and benchmarks.
Pooling the model--benchmark pairs yields a fitted slope above $0.90$ with $R^2=0.97$, while the mean accuracy decrease across benchmarks ranges from only $0.1$ to $3.4$ points across the four models.
These results indicate that question-blind descriptions preserve sufficient visual information to serve as an reliabe proxy for the perceptual capability of the evaluated VLMs.

\subsection{Bottlenecks Are Model–Benchmark Specific}
\label{para-headroom}

\score reveals broad benchmark-level trends, but also substantial variation across models on the same benchmark.
Fig.~\ref{fig:para_main}(b) shows representative results, while complete measurements across all 17 benchmarks and four models are provided in Appx.~\ref{app:para}.

At a coarse level, the measured headroom is consistent with the intended characteristics of several benchmarks.
For example, PerceptionBench exhibits greater perceptual than reasoning headroom, whereas MMMU generally exhibits greater reasoning headroom as expected. However, benchmark characteristics do not determine the model-specific bottleneck.
On MME, reasoning headroom is larger for both Qwen models, perceptual and reasoning headroom are comparable for Llama-4-Scout-17B, and perceptual headroom is substantially larger for Gemma-3-27B.
Thus, even on the same benchmark, different models can leave qualitatively different opportunities for improvement.

The mismatch can be even more pronounced for benchmarks conventionally viewed as reasoning-oriented.
On MathVista, perceptual headroom exceeds reasoning headroom for all four evaluated models despite the benchmark's emphasis on multimodal reasoning. This implies that a model's failures on a reasoning benchmark might not primarily reflect its insufficient reasoning capabilities.

Together, these results indicate that perception and reasoning headroom depend jointly on the model and benchmark, and bottlenecks should instead be diagnosed at the model--benchmark level.

\section{Understanding the Perception–Reasoning Asymmetry in Visual RL}

Having established that perception and reasoning headroom vary substantially across model--benchmark pairs, we next ask whether these diagnosed bottlenecks help explain why reinforcement learning produces uneven gains across visual tasks.

\subsection{RL Exploits Reasoning Headroom More Than Perceptual Headroom}
\label{subsec:rl-reasoning}

Reinforcement learning has produced substantial gains in language-model reasoning~\citep{shao2024deepseekmath,yu2025dapo}, yet improvements in visual perception remain more limited and inconsistent~\citep{xiao2025perceptionr1,li2026frankenstein,wu2026asymmetric}.
Prior work has observed that post-training improves reasoning more readily than perception, but has primarily characterized this asymmetry through manual error analysis or controlled interventions.
With \score, we can instead relate RL gains to model--benchmark-specific perception and reasoning headroom under a common measurement framework. To characterize the relative contribution of reasoning among recoverable failures, we define the directional reasoning bottleneck index

\begin{equation}
    \score_{\mathrm{BI}_{R}}^{M,B}
    =
    \frac{|R^{M,B}|}
    {|P^{M,B}\cup R^{M,B}|}.
    \label{eq:para-reasoning-bi}
\end{equation}
This quantity measures how broadly the reasoning intervention covers the union of failures recoverable through either intervention.
Because $P^{M,B}$ and $R^{M,B}$ may overlap, $\score_{\mathrm{BI}_{R}}$ should be interpreted as a directional coverage measure rather than as a decomposition of the error set.

\begin{wrapfigure}{r}{0.5\textwidth}
    \centering
    \includegraphics[width=\linewidth]{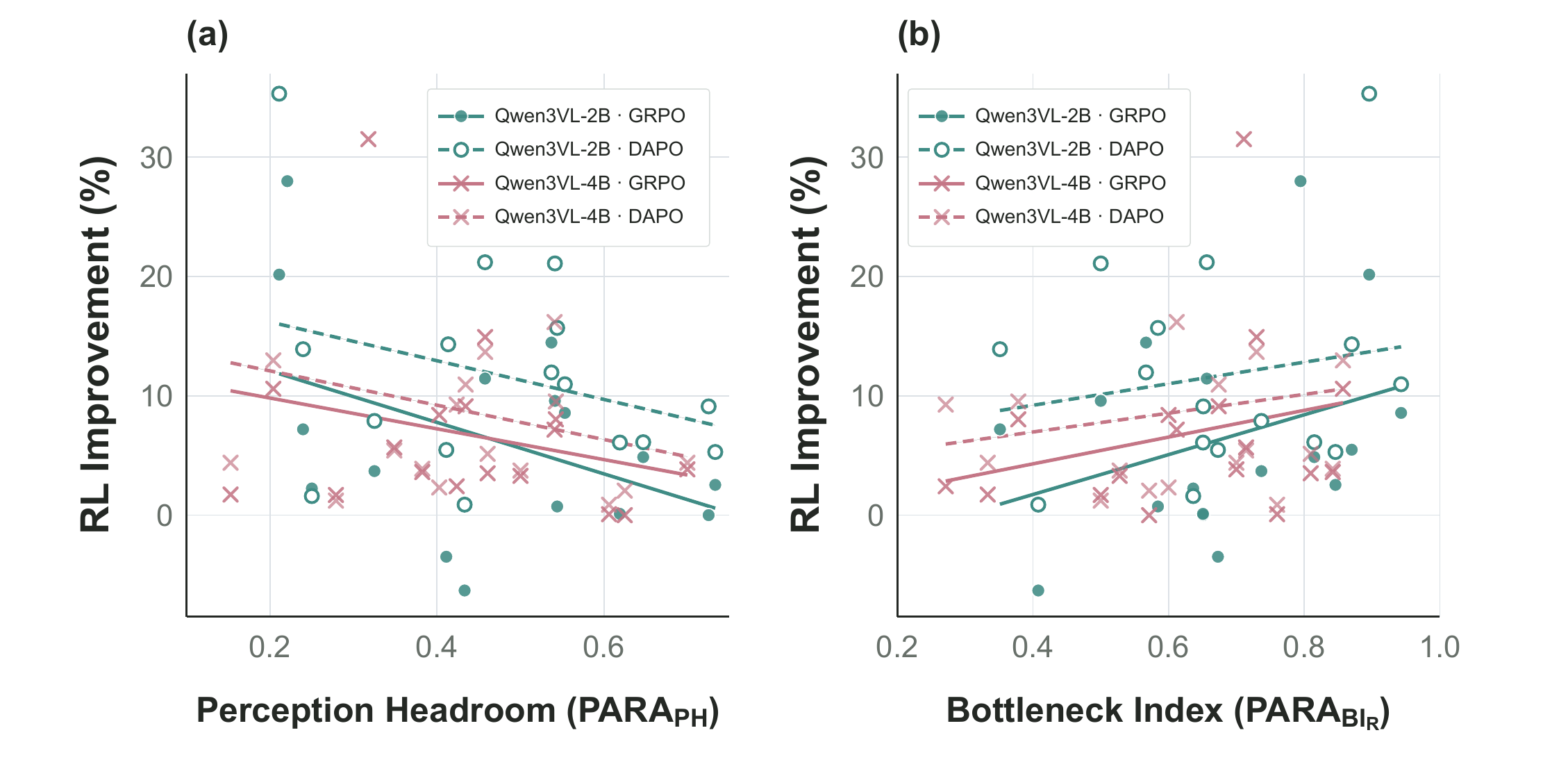}
    \vspace{-2\baselineskip}
    \caption{
    \textbf{Asymmetric RL gains.}
    Across models and algorithms , RL gains
    \textbf{(a)} decrease with perceptual headroom $\score_{\mathrm{PH}}$ and
    \textbf{(b)} increase with the reasoning bottleneck index $\score_{\mathrm{BI}_{R}}$.
    }
    \label{fig:grpo-vs-para}
\end{wrapfigure}

We compare the relative improvement of model after RL over its corresponding base model with two quantities measured on base model: perceptual headroom $\score_{\mathrm{PH}}$ and reasoning bottleneck index $\score_{\mathrm{BI}_{R}}$.
As in Fig.~\ref{fig:grpo-vs-para}, the same directional trends hold across both RL algorithms and model scales.
RL gains decrease as $\score_{\mathrm{PH}}$ increases, but increase with $\score_{\mathrm{BI}_{R}}$.
In other words, RL tends to yield smaller improvements when a larger fraction of the base model's failures is recoverable through stronger perception, while yielding larger improvements when reasoning intervention covers more of the recoverable failure set.

We interpret these relationships as cross-benchmark directional trends rather than predictions for individual tasks.
Nevertheless, their consistency across algorithms and models reveals a clear asymmetry: current RL objectives appear to exploit reasoning headroom more effectively than perceptual.

\subsection{Existing RL Objectives Indirectly Optimize Visual Attention}
\label{subsec:indirect}
Existing methods that target visual perception during RL primarily modify either the reward assigned to a rollout or the allocation of policy updates across output tokens~\citep{li2025visionsr1,huang2025spotlight}.
Although these approaches differ operationally, both ultimately modify the effective advantage used by the policy-gradient objective.
We therefore examine how such advantage-level signals propagate to the model's visual-attention pathway.

Given a rollout $\tau$ and output token $t$, let $\rho_t(\theta)$ denote the importance ratio and $\widetilde{A}_t(\tau)$ the effective advantage.
For simplicity, we write the GRPO objective as
\begin{equation}
    \mathcal{L}(\theta)
    =
    \mathbb{E}_{\tau,t}
    \left[
        \ell\!\left(
            \rho_t(\theta),
            \widetilde{A}_t(\tau)
        \right)
    \right].
    \label{eq:advantage-interface}
\end{equation}
Reward-level methods modify the value of $\widetilde{A}_t$, while token-level methods redistribute it across output tokens.
In either case, $\widetilde{A}_t$ is treated as fixed during a policy update.

Within a VLM attention block, let visual contribution to the representation of generated token $t$ be
\begin{equation}
    h_t^{\mathrm{vis}}
    =
    \sum_{j\in\mathcal{V}}
    a_{t,j}v_j,
\end{equation}
where $v_j$ denotes the value representation of visual token $j$ and $a_{t,j}$ the attention from output token $t$ to visual token $j$.
Collecting the visual representations and attention weights as $h^{\mathrm{vis}}$ and $a$, respectively, the loss along this pathway has the composition
$\mathcal{L}(z(h^{\mathrm{vis}}(a));\widetilde{A})$.
By the chain rule,
\begin{equation}
    \nabla_a \mathcal{L}
    =
    \left(
        \frac{\partial h^{\mathrm{vis}}}{\partial a}
    \right)^{\!\top}
    \left(
        \frac{\partial z}{\partial h^{\mathrm{vis}}}
    \right)^{\!\top}
    \nabla_z \mathcal{L}.
    \label{eq:advantage-indirect}
\end{equation}
Eq.~\ref{eq:advantage-indirect} makes the optimization path to visual attention explicit.
The term $\nabla_z\mathcal{L}$ determines the update to the output logits,
$\partial z/\partial h^{\mathrm{vis}}$ propagates this signal to the visual contribution,
and $\partial h^{\mathrm{vis}}/\partial a$ further propagates it to the attention allocation.
Thus, reward- and token-level perceptual signals can influence visual attention, but only through their downstream effect on the output logits.
The objective contains no separate term that directly evaluates or supervises the visual-attention distribution itself.

We examine this distinction empirically using Vision-SR1~\citep{li2025visionsr1} and VPPO~\citep{huang2025spotlight} as representative reward-level and token-level methods, respectively.
Vision-SR1 rescales the rollout advantage using a perceptual reward weight $W_p$, whereas VPPO redistributes token-level advantages through a trajectory weight $\alpha(\tau)$ and a visual mask $m_t(k)$.
Both therefore alter the strength or allocation of $\widetilde{A}_t$ without directly supervising visual attention.

As shown in Fig.~\ref{fig:supervision-sweep}(b) and~\ref{fig:supervision-sweep}(c), strengthening these perceptual modifications, either by increasing $W_p$ for Vision-SR1 or decreasing the mask-token ratio $k$ for VPPO, does not systematically produce larger gains on benchmarks with greater perceptual headroom $\score_{\mathrm{PH}}$.
Meanwile, increasing the strength of either intervention does not consistently rotate the fitted relationship toward a more positive slope.
These results do not imply that advantage-level methods fail to improve overall performance; rather, they suggest that increasing their perceptual signal does not systematically translate into greater benefit on tasks with more available perceptual headroom.

\subsection{Visual Attention Provides a Direct Optimization Interface}
\label{subsec:direct}

The preceding analysis suggests that directly supervising visual attention provides a more immediate optimization path to the model's visual pathway.
Suppose we augment the RL objective with an auxiliary regularizer $\mathcal{R}(a)$ on the attention distribution, weighted by $\lambda$.
The resulting gradient with respect to attention becomes
\begin{equation}
    \nabla_a \mathcal{L}^{a}
    =
    \left(
        \frac{\partial h^{\mathrm{vis}}}{\partial a}
    \right)^{\!\top}
    \left(
        \frac{\partial z}{\partial h^{\mathrm{vis}}}
    \right)^{\!\top}
    \nabla_z \mathcal{L}
    +
    \lambda \nabla_a \mathcal{R}(a).
    \label{eq:attention-direct}
\end{equation}
Unlike advantage-level supervision, the additional term
$\lambda\nabla_a\mathcal{R}(a)$
acts directly on the attention distribution rather than reaching it only through the output logits.

We examine this direct pathway using On-Policy Attention Distillation (OPAD) from RAL~\citep{li2026ral}.
Let $p_t^{\theta}$ denote the current visual-attention distribution for token $t$, and let $p_t^{\mathrm{old}}$ denote the corresponding distribution recorded during rollout generation.
OPAD uses the regularizer
\begin{equation}
    \mathcal{R}(a)
    =
    \mathbb{E}_{\tau,t}
    \left[
        \widetilde{A}_t(\tau)
        D_{\mathrm{JS}}
        \left(
            p_t^{\theta}
            \,\middle\|\,
            p_t^{\mathrm{old}}
        \right)
    \right],
    \label{eq:opad-attention}
\end{equation}
where $D_{\mathrm{JS}}$ denotes the Jensen--Shannon divergence.
Positive advantages encourage preservation of attention patterns associated with successful rollouts, whereas negative advantages push the current policy away from attention patterns associated with unsuccessful rollouts.

As shown in Fig.~\ref{fig:supervision-sweep}(d), increasing the coefficient $\lambda$ systematically changes the relationship between performance gain and perceptual headroom: the fitted slope moves from negative toward approximately flat, reducing the performance disadvantage associated with larger $\score_{\mathrm{PH}}$. Together, the theoretical and empirical results suggest that visual attention provides a more direct interface for addressing the perceptual headroom identified by \score.
\section{GRAD: Group-Relative Attention Distillation for Perception}
\label{sec:method}

Sec.~\ref{subsec:direct} identifies visual attention as a direct optimization interface for addressing perceptual bottlenecks.
However, direct attention supervision alone does not guarantee a reliable update direction.
Existing attention-level objectives remain coupled to rollout outcomes: they preserve the old-policy attention of successful rollouts while pushing unsuccessful rollouts away from their previous attention patterns.
This coupling is ambiguous because an outcome reward jointly reflects perception and reasoning.
A rollout may fail because of incorrect reasoning despite attending to the relevant visual evidence, in which case pushing away from its attention pattern could penalize valid perception.

Consistent with this limitation, while OPAD changes the perception-dependent trend in Fig.~\ref{fig:supervision-sweep}(d), it still yields only moderate overall gains relative to other methods in Fig.~\ref{fig:supervision-sweep}(b)(c).
These observations suggest that attention provides the appropriate optimization interface, but effective perceptual supervision requires a more careful objective design.

We therefore propose \model, \emph{Group-Relative Attention Distillation}.
Rather than using each rollout's own old-policy attention as its reference, \model constructs a positive attention target from the best sibling within the same on-policy rollout group.
For a rollout group $g$ generated from the same image and prompt, we select
\begin{equation}
    b(g)
    =
    \arg\max_{j\in g_{\mathrm v}} r_j,
    \label{eq:grad-best-sibling}
\end{equation}
where $r_j$ is the reward of rollout $j$ and $g_{\mathrm v}$ denotes the subset of rollouts with valid attention distributions.
When multiple siblings attain the same maximal reward, we select the rollout with the shortest trajectory. We then treat negative-advantage rollouts as students and distill the best sibling's visual-attention pattern into them.
For rollout $i$, we define the adaptive supervision weight
\begin{equation}
    w_i = \max(-A_i,0),
\end{equation}
such that supervision is activated only when $A_i<0$ and becomes stronger as the advantage decreases.
Groups with identical rewards are therefore naturally excluded because they contain no negative-advantage students.
Importantly, $w_i$ is always nonnegative, so minimizing the distillation objective consistently pulls a student's attention toward the best sibling rather than pushing it away from its own previous attention.
This converts outcome feedback into a positive attention target without requiring an external teacher or additional annotation.

To focus the auxiliary objective specifically on perception, we perform alignment only over visual tokens.
For rollout $i$, let $\bar p_i^{\theta}$ denote its visual-attention distribution, obtained by renormalizing attention over visual tokens and pooling across response tokens and attention heads.
Because sibling rollouts share the same visual input, their resulting visual-attention distributions are directly comparable even when their generated responses have different lengths.
Let $\widehat p_g$ denote the corresponding visual-attention distribution of the best sibling $b(g)$.
The \model objective is
\begin{equation}
    \mathcal L_{\model}
    =
    \frac{1}{|\mathcal I|}
    \sum_{(g,i)\in\mathcal I}
    w_i
    D_{\mathrm{JS}}
    \left(
        \bar p_i^{\theta}
        \,\middle\|\,
        \widehat p_g
    \right),
    \label{eq:grad-attention-loss}
\end{equation}
where $\mathcal I$ denotes the set of valid negative-advantage students.

The complete training objective combines standard GRPO objective with outcome reward and group-relative attention distillation tailored for mitigating perception bottleneck:
\begin{equation}
    \mathcal L_{\mathrm{total}}
    =
    \mathcal L_{\mathrm{GRPO}}
    +
    \lambda \mathcal L_{\model},
    \label{eq:grad-total-loss}
\end{equation}
where $\lambda$ controls the strength of attention supervision.
By constructing perceptual supervision directly from on-policy sibling rollouts, \model requires no external teacher, additional annotations, extra rollouts, or inference-time modifications.
\section{Experiments}
\label{sec:experiments}

\subsection{Experimental Setup}

\paragraph{Dataset and Benchmarks.}
Following VisionSR1~\citep{li2025visionsr1}, we use \texttt{Vision-SR1-47K} as the training dataset for RL.
We evaluate on 7 visual benchmarks, including MME~\citep{fu2023mme}, MMMU~\citep{yue2024mmmu}, MMMU-Pro~\citep{yue2025mmmupro}, MATH-Vision~\citep{wang2024mathvision}, PerceptionBench~\citep{lin2026perceptionbench}, ScienceQA~\citep{lu2022scienceqa}, and VisNumBench~\citep{weng2025visnumbench}. We report MME on its original scale and normalize it to a $100$-point scale only when computing the average score. We follow the official evaluation protocol and metric for each benchmark.

\begin{table}[t]
\centering
\scriptsize
\vspace{-1.5\baselineskip}
\caption{\textbf{Results across seven visual benchmarks.} MME is reported on its native scale, while Avg. rescales it with $1/28$. The best and second-best results are shown in \textbf{bold} and \underline{underline}, respectively.}
\label{tab:main-results}
\resizebox{\linewidth}{!}{%
\begin{tabular}{clrrrrrrrr}
\toprule
& Method & MME & MMMU & MMMU-Pro & MathVis & PercepBench & ScienceQA & VisNum. & Avg. \\
\midrule
\multirow{6}{*}{\rotatebox[origin=c]{90}{\textit{Qwen3-VL-2B}}}
& Base & 2044.0 & 46.6 & 30.8 & 13.5 & 22.7 & 86.1 & 35.1 & 44.0 \\
& GRPO & 2217.6 & 49.1 & 34.3 & 18.1 & 21.3 & \underline{88.2} & 35.3 & 46.5 \\
& Vision-SR1 & 2223.2 & 48.3 & \underline{34.8} & \underline{20.7} & \underline{23.7} & 87.5 & \textbf{41.1} & \underline{47.9} \\
& VPPO & 2212.0 & \underline{50.8} & 33.8 & 19.1 & 22.8 & \underline{88.2} & 37.6 & 47.3 \\
& OPAD & \underline{2256.8} & 49.6 & 32.9 & 16.1 & 20.3 & 87.1 & 36.3 & 46.1 \\
\cmidrule(lr){2-10}
& Ours & \textbf{2290.4} & \textbf{52.1} & \textbf{35.1} & \textbf{21.1} & \textbf{24.4} & \textbf{90.3} & \underline{40.8} & \textbf{49.4} \\
\midrule
\multirow{6}{*}{\rotatebox[origin=c]{90}{\textit{Qwen3-VL-4B}}}
& Base & 2248.4 & 60.1 & 45.6 & 30.3 & 24.8 & 91.7 & 49.7 & 54.6 \\
& GRPO & 2240.0 & 62.1 & 48.4 & 37.1 & 25.7 & 93.9 & 52.1 & 57.0 \\
& Vision-SR1 & \underline{2332.4} & 60.6 & 47.9 & \underline{38.2} & \textbf{29.1} & 94.3 & \underline{53.7} & \underline{58.2} \\
& VPPO & 2315.6 & 62.9 & \underline{48.6} & 35.5 & 25.9 & 94.3 & 52.9 & 57.5 \\
& OPAD & 2279.2 & \underline{63.1} & 48.0 & 37.8 & 25.1 & \underline{94.4} & 51.4 & 57.3 \\
\cmidrule(lr){2-10}
& Ours & \textbf{2349.2} & \textbf{64.3} & \textbf{49.8} & \textbf{40.5} & \underline{27.9} & \textbf{95.0} & \textbf{55.0} & \textbf{59.5} \\
\bottomrule
\vspace{-0.5\baselineskip}
\end{tabular}}
\end{table}

\paragraph{Baselines.}
Beyond GRPO~\citep{shao2024deepseekmath}, we compare against baselines spanning diverse visual supervision. VisionSR1~\citep{li2025visionsr1} introduces a perception reward, VPPO~\citep{huang2025spotlight} reallocates policy updates across response tokens, and OPAD~\citep{li2026ral} directly regularizes visual attention with old policy. We additionally use DAPO~\citep{yu2025dapo} to test whether our method complements other RL algorithms. 

\paragraph{Implementation Details.} We use \emph{Qwen3-VL-2B-Instruct} and \emph{Qwen3-VL-4B-Instruct}~\citep{bai2025qwen3vl} and perform full fine-tuning on four NVIDIA H200 GPUs with random seed $2026$. We use a global batch size of $128$ and sample eight rollouts per prompt. Details are provided in Appx.~\ref{app:experiments}.

\subsection{Experimental Results}
\paragraph{Main Results.}
In Tab.~\ref{tab:main-results}, we compare \model with baselines across seven visual benchmarks. \model achieves the highest average performance for both models, outperforming GRPO by $2.9$ and $2.5$ points, respectively.
On MME, \model improves over GRPO by $72.8$ and $109.2$ points, respectively.
OPAD performs comparably to GRPO overall, whereas \model improves upon OPAD by $3.3$ and $2.2$ points, highlighting the benefit of our group-relative design.
Moreover, although VPPO and Vision-SR1 improve through objectives orthogonal to ours, \model exceeds both by at least $1.5$ points at 2B and $1.3$ points at 4B on average, demonstrating our overall effectiveness.

\begin{table}[t]
\centering
\scriptsize
\vspace{-0.5\baselineskip}
\caption{\textbf{Complementarity with existing methods.} Each entry reports the performance integrating \model to corresponding baseline on Qwen3-VL-2B. The largest gain at each benchmark is \textbf{bolded}.}
\label{tab:method-combinations}
\resizebox{1.0\linewidth}{!}{%
\begin{tabular}{lrrrrrrrr}
\toprule
Method & MME & MMMU & MMMU-Pro & MathVis & PercepBench & ScienceQA & VisNum. & Avg. \\
\midrule
Vision-SR1 w/ Ours & +44.8 & \textbf{+2.8} & +0.2 & \textbf{+4.3} & +0.8 & \textbf{+2.4} & \textbf{+2.6} & \textbf{+2.1} \\
VPPO w/ Ours & +19.6 & +0.5 & +0.5 & +1.3 & -0.9 & +1.6 & +1.3 & +0.7 \\
DAPO w/ Ours & \textbf{+123.2} & -0.2 & \textbf{+0.6} & +2.3 & \textbf{+1.7} & +2.3 & +1.8 & +1.9 \\
\bottomrule
\vspace{-0.5\baselineskip}
\end{tabular}}
\end{table}

\begin{figure*}[t]
\centering
\includegraphics[width=\textwidth]{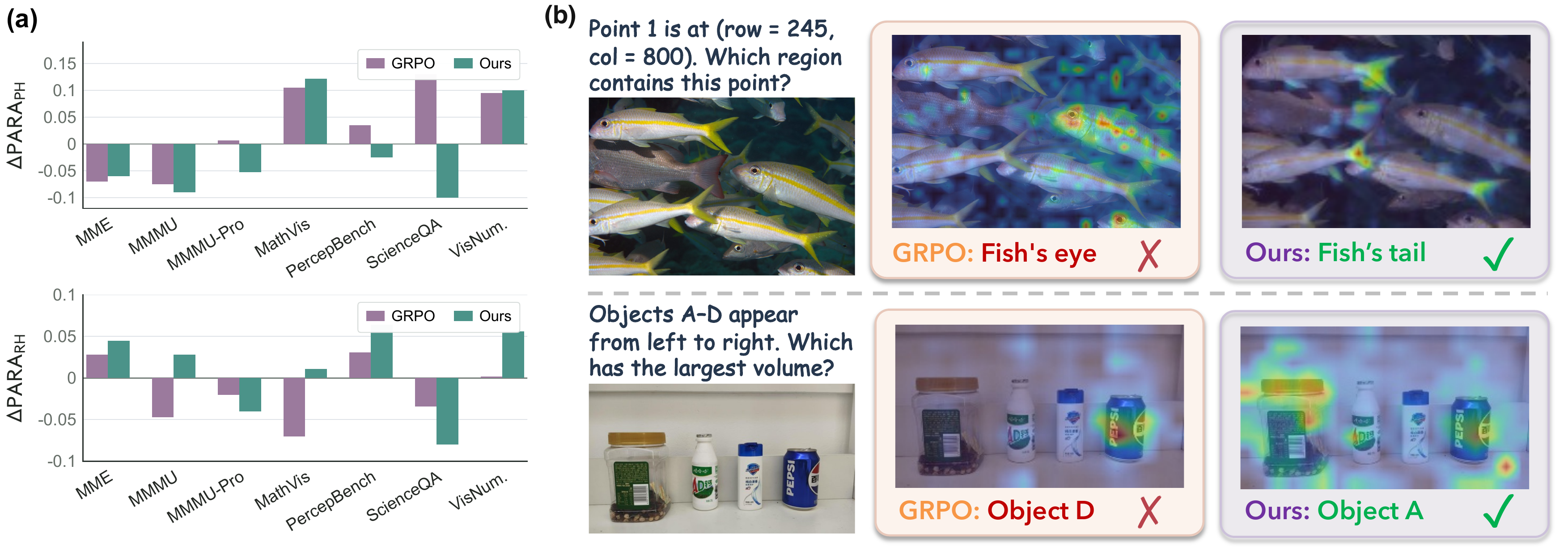}
\caption{\textbf{(a) Changes in perception and reasoning headroom relative to Qwen3-VL-4B.} \model reduces perceptual headroom consistently over GRPO while making more residual errors recoverable through reasoning intervention. \textbf{(b) Visualization of visual attention map.} \model focuses on the correct visual evidence while GRPO fails.}
\label{fig:para-attention}
\end{figure*}

\paragraph{Analysis of Perceptual Improvement.}
In Fig.~\ref{fig:para-attention}(b), we further visualize inference attention in representative cases from PerceptionBench and VisNumBench. In the upper case, \model shifts attention from the fish's eye to its tail, and in the lower case, it focuses on Object A rather than the irrelevant Object D, enabling the correct volume comparison. These cases provide qualitative evidence that \model improves perception by direct visual attention supervision.

Given \model aiming to improve perception, we further investigate how it changes the residual headroom.
As shown in Fig.~\ref{fig:para-attention}(a), \model reduces $\score_{\mathrm{PH}}$ on five benchmarks, while GRPO reduces only two.
On MMMU-Pro, PerceptionBench, and ScienceQA, \model reverses the increase into a reduction.
Across the remaining four benchmarks, \model further increases $\score_{\mathrm{RH}}$ by a substantially larger margin, indicating that stronger perception preserves more usable visual evidence and allows the same reasoning intervention to recover a larger fraction of failures.
Together, the reduction in perceptual headroom and increase in reasoning headroom provide complementary evidence that \model directly strengthens perception during RL.

\paragraph{Complementarity with Existing Methods.}
Furthermore, we evaluate whether \model can be combined with existing methods. As shown in Tab.~\ref{tab:method-combinations}, integrating our objective improves the average scores of Vision-SR1, VPPO, and DAPO by $2.1$, $0.7$, and $1.9$ points, respectively. The combinations improve all seven benchmarks for Vision-SR1 and six of seven for both VPPO and DAPO. These gains demonstrate that our attention-level objective complements methods at the reward, token, and policy levels, and can be readily integrated with existing methods.

\subsection{Ablation Studies}
\label{sec:ablation}

\paragraph{Method Components.}
Tab.~\ref{tab:component-ablation} examines each component of \model.
On Qwen3-VL-2B, removing the perception-only restriction lowers the average score by $1.5$ points, yet the resulting variant remains above GRPO, indicating that our objective supplies the perceptual signal missing from plain RL.
Further removing best sibling supervision, randomly choosing a target sibling instead, reduces the score to $45.5$, confirming the importance of a reliable group-relative target.
In addition, we evaluate the performance of keeping only the tokens attribute to top 50\% attention in the best sibling. Adding this constraints has only modest effects on most benchmarks and lowers the average by $1.1$ points. 
Together, these results support our perceptual and group-relative attention design.

\begin{table}[t]
\centering
\scriptsize
\caption{\textbf{Ablation study results.} Both components work jointly to achieve better performance.}
\label{tab:component-ablation}
\resizebox{1.0\linewidth}{!}{%
\begin{tabular}{lrrrrrrrr}
\toprule
Method & MME & MMMU & MMMU-Pro & MathVis & PercepBench & ScienceQA & VisNum. & Avg. \\
\midrule
Ours & \textbf{2290.4} & \underline{52.1} & \textbf{35.1} & \underline{21.1} & \textbf{24.4} & \textbf{90.3} & \textbf{40.8} & \textbf{49.4} \\
- w/ Attention Mass. & 2265.2 & 50.7 & \underline{34.6} & \textbf{21.2} & \underline{21.7} & 88.8 & \underline{40.2} & \underline{48.3} \\
- w/o Percep. Only & 2228.8 & \textbf{52.7} & 34.2 & 19.1 & 21.2 & \underline{90.1} & 38.2 & 47.9 \\
- w/o Best Sibling & \underline{2284.8} & 50.7 & 32.4 & 13.5 & 19.1 & 87.3 & 33.6 & 45.5 \\
\bottomrule
\end{tabular}}
\end{table}

\paragraph{Hyper-parameters.}
Following \citet{li2026ral}, we set $\lambda=1.0$ and supervise only the last decoder layer throughout our experiments. Setting $\lambda=0.2$ or $\lambda=5.0$ lowers the average score by about $2$ points, although both configurations remain better than GRPO.
Extending supervision to the top 4 or 8 layers also underperforms the last layer setting, while applying it to the top 16 or all layers leads to substantially larger performance drops.
These denser targets appear more difficult to optimize, suggesting that effective multi-layer supervision requires a dedicated design. Detailed results of ablation study are provided in Appx.~\ref{app:experiment-ablation}.

\section{Conclusion}
In this work, we examine how much perceptual and reasoning headroom remains for VLMs across visual benchmarks, and how these two forms of headroom relate to the gains from RL post-training.
We introduce \score, a quantitative measure that estimates, for each model--benchmark pair, how many current errors can be recovered with better visual information versus better reasoning.
Across diverse models and benchmarks, \score reveals substantial model-specific variation and a consistent asymmetry in RL: gains decrease with greater perceptual headroom but increase as reasoning accounts for more recoverable errors.
We further show that reward- and token-level objectives affect visual attention only indirectly, motivating more direct attention-level supervision.
Accordingly, we propose \model, a group-relative attention distillation objective that consistently improves performance and remains complementary to existing RL methods.
Together, our results connect model-specific bottleneck measurement with targeted optimization for improving visual RL.

\bibliography{references}
\bibliographystyle{plainnat}

\newpage
\appendix
\section{\score Results}
\label{app:para}

\subsection{Full Benchmark Results}
Tab.~\ref{tab:para-main} reports the full \score results for all $17$ benchmarks across the four models evaluated in Sec.~\ref{subsec:para}, including Qwen3-VL-2B and Qwen3-VL-4B~\citep{bai2025qwen3vl}, Llama-4-Scout-17B~\citep{meta2025llama4}, and Gemma-3-27B~\citep{gemmateam2025gemma3}.

\begin{table*}[h]
\centering
\caption{\textbf{Full \score results across models and benchmarks.} PH and RH denote perception and reasoning headroom, and $\mathrm{BI}_{P}$ and $\mathrm{BI}_{R}$ denote their directional bottleneck indices. The complete results further support the benchmark-level trends alongside distinct patterns across models.}
\label{tab:para-main}
\resizebox{\textwidth}{!}{%
\begin{tabular}{lrrrrrrrrrrrrrrrr}
\toprule
Benchmark & \multicolumn{4}{c}{Qwen3-VL-2B} & \multicolumn{4}{c}{Qwen3-VL-4B} & \multicolumn{4}{c}{Llama-4-Scout-17B} & \multicolumn{4}{c}{Gemma-3-27B} \\
\cmidrule(lr){2-5} \cmidrule(lr){6-9} \cmidrule(lr){10-13} \cmidrule(lr){14-17}
 & $\mathrm{PH}$ & $\mathrm{RH}$ & $\mathrm{BI}_{P}$ & $\mathrm{BI}_{R}$ & $\mathrm{PH}$ & $\mathrm{RH}$ & $\mathrm{BI}_{P}$ & $\mathrm{BI}_{R}$ & $\mathrm{PH}$ & $\mathrm{RH}$ & $\mathrm{BI}_{P}$ & $\mathrm{BI}_{R}$ & $\mathrm{PH}$ & $\mathrm{RH}$ & $\mathrm{BI}_{P}$ & $\mathrm{BI}_{R}$ \\
\midrule
BLINK & 0.54 & 0.35 & 0.77 & 0.50 & 0.54 & 0.24 & 0.86 & 0.38 & 0.62 & 0.26 & 0.96 & 0.40 & 0.61 & 0.39 & 0.85 & 0.54 \\
VisNumBench & 0.54 & 0.42 & 0.76 & 0.58 & 0.54 & 0.42 & 0.79 & 0.61 & 0.53 & 0.46 & 0.74 & 0.64 & 0.60 & 0.44 & 0.81 & 0.59 \\
RealWorldQA & 0.54 & 0.31 & 0.97 & 0.57 & 0.46 & 0.46 & 0.73 & 0.73 & 0.50 & 0.33 & 0.86 & 0.57 & 0.67 & 0.42 & 0.98 & 0.61 \\
PerceptionBench & 0.43 & 0.20 & 0.89 & 0.41 & 0.42 & 0.13 & 0.87 & 0.27 & 0.38 & 0.19 & 0.86 & 0.42 & 0.35 & 0.16 & 0.80 & 0.37 \\
MMStar & 0.62 & 0.49 & 0.83 & 0.65 & 0.50 & 0.32 & 0.83 & 0.53 & 0.47 & 0.31 & 0.81 & 0.53 & 0.58 & 0.36 & 0.88 & 0.56 \\
ChartQA & 0.33 & 0.35 & 0.68 & 0.74 & 0.35 & 0.35 & 0.71 & 0.71 & 0.23 & 0.13 & 1.00 & 0.57 & 0.46 & 0.35 & 0.91 & 0.70 \\
HallusionBench & 0.73 & 0.55 & 0.86 & 0.65 & 0.62 & 0.42 & 0.86 & 0.57 & 0.58 & 0.30 & 0.93 & 0.48 & 0.68 & 0.39 & 0.94 & 0.53 \\
MME & 0.55 & 0.70 & 0.74 & 0.94 & 0.38 & 0.47 & 0.68 & 0.84 & 0.55 & 0.53 & 0.88 & 0.83 & 0.79 & 0.51 & 0.98 & 0.63 \\
MMBench & 0.65 & 0.65 & 0.81 & 0.81 & 0.61 & 0.58 & 0.80 & 0.76 & 0.48 & 0.36 & 0.92 & 0.69 & 0.46 & 0.38 & 0.85 & 0.69 \\
ScienceQA & 0.73 & 0.73 & 0.85 & 0.85 & 0.70 & 0.70 & 0.70 & 0.70 & 0.90 & 0.60 & 1.00 & 0.60 & 0.46 & 0.54 & 0.60 & 0.70 \\
MM-Vet & 0.25 & 0.17 & 0.91 & 0.64 & 0.28 & 0.16 & 0.85 & 0.50 & 0.33 & 0.22 & 0.78 & 0.53 & 0.30 & 0.25 & 0.84 & 0.71 \\
MMMU & 0.41 & 0.64 & 0.56 & 0.87 & 0.46 & 0.57 & 0.65 & 0.81 & 0.51 & 0.52 & 0.75 & 0.77 & 0.47 & 0.54 & 0.69 & 0.78 \\
MMMU-Pro & 0.46 & 0.45 & 0.67 & 0.66 & 0.43 & 0.41 & 0.72 & 0.67 & 0.32 & 0.40 & 0.58 & 0.73 & 0.31 & 0.45 & 0.52 & 0.77 \\
LogicVista & 0.41 & 0.41 & 0.67 & 0.67 & 0.40 & 0.31 & 0.78 & 0.60 & 0.55 & 0.39 & 0.75 & 0.54 & 0.60 & 0.23 & 0.86 & 0.33 \\
MathVista & 0.24 & 0.09 & 0.92 & 0.35 & 0.15 & 0.06 & 0.83 & 0.33 & 0.22 & 0.08 & 0.94 & 0.32 & 0.18 & 0.09 & 0.90 & 0.43 \\
MathVerse & 0.21 & 0.34 & 0.56 & 0.90 & 0.20 & 0.29 & 0.60 & 0.86 & 0.19 & 0.21 & 0.59 & 0.66 & 0.19 & 0.23 & 0.63 & 0.77 \\
MATH-Vision & 0.22 & 0.43 & 0.41 & 0.79 & 0.32 & 0.30 & 0.76 & 0.71 & 0.27 & 0.39 & 0.48 & 0.71 & 0.41 & 0.34 & 0.75 & 0.61 \\
\midrule
\textit{Mean} & 0.46 & 0.43 & 0.76 & 0.68 & 0.43 & 0.36 & 0.77 & 0.62 & 0.45 & 0.33 & 0.81 & 0.59 & 0.48 & 0.36 & 0.81 & 0.61 \\
\bottomrule
\end{tabular}}
\end{table*}

\subsection{Prompt Templates and Evaluation Pipeline}

We acknowledge that evaluating every example in each benchmark would yield more accurate measurements. Due to computational resource constraints, in this work, we use $200$ examples per benchmark as an estimate and apply exact-match evaluation to the boxed final answer. The nine natively multiple-choice benchmarks use the multiple-choice answer format.
We apply the same answer instruction to the base condition and all corresponding intervention conditions. The self-description validation uses the same instruction for both inputs so that the image and description conditions differ only in their visual input modality.

Tab.~\ref{tab:para-prompts} summarizes the templates used in our evaluation. The oracle and target model
descriptions use the same question-blind instruction, and no condition uses a separate system prompt.

\begin{table}[h]
\centering
\caption{\textbf{Prompt templates used for \score.}}
\label{tab:para-prompts}
\small
\begin{tabular}{@{}p{0.15\linewidth}p{0.82\linewidth}@{}}
\toprule
Template & Prompt \\
\midrule
Question\newline answering & You FIRST think about the reasoning process as an internal monologue and then provide
the final answer. The reasoning process MUST BE enclosed within \texttt{<think> </think>} tags. The final answer
MUST BE put in \texttt{\textbackslash boxed\{\}}. For multiple-choice questions, put the letter of the correct
option in \texttt{\textbackslash boxed\{\}}, e.g., \texttt{\textbackslash boxed\{A\}}. \\
\midrule
Question-blind description & Describe this image in exhaustive detail for someone who cannot see it. Report
every object, text, number, axis label, chart value, spatial relationship, color, and fine visual detail you can
discern. Do NOT answer any question or guess what is being asked---only describe what is literally visible. \\
\midrule
Description-conditioned answer & \texttt{Visual description:} [visual description]\newline
\texttt{Question:} [question]\newline [Question answering instruction] \\
\bottomrule
\end{tabular}
\end{table}

Tab.~\ref{tab:self-description-examples} presents representative question-blind self-descriptions from the four benchmarks highlighted in the main text. These examples show that the descriptions capture visible content in detail without introducing question-conditioned reasoning or answer information. Tabs.~\ref{tab:oracle-description-examples-claude} and~\ref{tab:oracle-description-examples-gpt} additionally show descriptions from Claude Opus 4.8 and GPT-5.6-Luna on the same benchmarks using distinct examples. Tab.~\ref{tab:perception-recovery-example} presents a perception-recovery case in which the GPT-5.6-Luna description changes Qwen3-VL-4B's answer from A to the correct answer C.

\begin{table*}[h]
\centering
\caption{\textbf{Representative examples of question-blind descriptions.} Each description is generated by Qwen3-VL-4B without seeing the question.}
\label{tab:self-description-examples}
\setlength{\tabcolsep}{6pt}
\renewcommand{\arraystretch}{1.12}
\scriptsize
\begin{tabular}{@{}p{0.27\textwidth}p{0.70\textwidth}@{}}
\toprule
Visual Input & Question-blind Description \\
\midrule
\textbf{PerceptionBench}\par\medskip
\centering\includegraphics[width=0.78\linewidth]{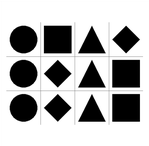}
&
The image is a $3\!\times\!4$ grid composed of 12 distinct black shapes on a plain white background. Each cell contains a single solid black shape. The top row contains a circle, square, upward-pointing triangle, and diamond. The middle row contains a circle, diamond, upward-pointing triangle, and square. The bottom row repeats the middle row: a circle, diamond, upward-pointing triangle, and square. Each shape is centered within its cell, and the shapes do not overlap. \\
\midrule
\textbf{MMMU}\par\medskip
\centering\includegraphics[width=0.94\linewidth]{figures/self_description_examples/mmmu_97.png}
&
The image displays a table with two rows and six columns. The top row is labeled $x$, followed by the values $-1$, $-0.6$, $-0.2$, $0.2$, and $0.6$. The bottom row is labeled $f'(x)$, followed by the corresponding values $1$, $2$, $-0.5$, $-1.5$, and $1.2$. The values are centered in their cells, with black text and lines on a white background. \\
\midrule
\textbf{MathVista}\par\medskip
\centering\includegraphics[width=0.75\linewidth]{figures/self_description_examples/mathvista_121.png}
&
The image displays a single solid-colored rectangle centered on a plain white background. The rectangle is filled with a uniform muted purple color and is wider than it is tall. The top and bottom sides are each labeled ``3 cm,'' while the left and right sides are each labeled ``2 cm.'' The labels are positioned outside the corresponding sides, and the rectangle has sharp right-angled corners. \\
\midrule
\textbf{MME}\par\medskip
\centering\includegraphics[width=0.68\linewidth]{figures/self_description_examples/mme_93.png}
&
The image displays a broom resting on a green grassy surface. It has a long, slender, light-colored wooden handle with a bright yellow rounded cap. The handle connects to a wide, dark-blue rectangular head with a raised central joint. Densely packed, light-colored straw bristles fan outward below the head and appear slightly worn at their tips. The grass contains scattered brown patches, and a partially visible dark object with a small red or orange area appears near the upper-right edge. \\
\bottomrule
\end{tabular}
\end{table*}

\begin{table*}[h]
\centering
\caption{\textbf{Representative question-blind descriptions from Claude Opus 4.8.}}
\label{tab:oracle-description-examples-claude}
\setlength{\tabcolsep}{6pt}
\renewcommand{\arraystretch}{1.12}
\scriptsize
\begin{tabular}{@{}p{0.27\textwidth}p{0.70\textwidth}@{}}
\toprule
Visual Input & Claude Opus 4.8 Description \\
\midrule
\textbf{PerceptionBench}\par\medskip
\centering\includegraphics[width=0.78\linewidth]{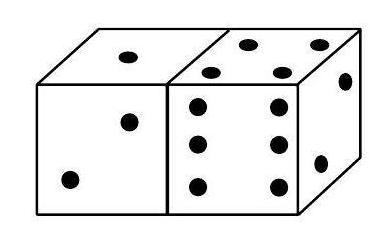}
&
The image is a black-and-white line drawing of two dice positioned side by side. The left die shows a front face with two diagonal dots and a top face with one dot. The right die shows a front face with six dots, a top face with three dots, and a right face with two dots. The dice touch along their adjacent edges on a plain white background. \\
\midrule
\textbf{MMMU}\par\medskip
\centering\includegraphics[width=0.94\linewidth]{figures/oracle_description_examples/claude_mmmu_52.png}
&
The image shows a table titled ``Cash Flows (\$)'' in white text on a blue banner. Its columns are Project and $C_0$ through $C_5$. Project A contains $-1{,}000$, $+1{,}000$, and four zeros. Project B contains $-2{,}000$, $+1{,}000$, $+1{,}000$, $+4{,}000$, $+1{,}000$, and $+1{,}000$. Project C contains $-3{,}000$, $+1{,}000$, $+1{,}000$, 0, $+1{,}000$, and $+1{,}000$. \\
\midrule
\textbf{MathVista}\par\medskip
\centering\includegraphics[width=0.75\linewidth]{figures/oracle_description_examples/claude_mathvista_21.png}
&
The image is a horizontal bar chart titled ``Title.'' Three medium-blue bars correspond to ``bull,'' ``wish,'' and ``heap.'' The x-axis is labeled ``Values'' and runs from 0 to 10 in increments of 2. The bars reach approximately 4, 9, and 1, respectively. The white plotting area contains faint grid lines and a rectangular border. \\
\midrule
\textbf{MME}\par\medskip
\centering\includegraphics[width=0.68\linewidth]{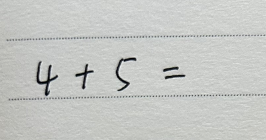}
&
The image shows off-white ruled paper with two gray dashed horizontal guide lines. Centered between them is the handwritten expression ``4 + 5 ='' in dark ink. The handwriting is informal, and the space to the right of the equals sign is blank. No other text, numbers, or objects are visible. \\
\bottomrule
\end{tabular}
\end{table*}

\begin{table*}[h]
\centering
\caption{\textbf{Representative question-blind descriptions from GPT-5.6-Luna.}}
\label{tab:oracle-description-examples-gpt}
\setlength{\tabcolsep}{6pt}
\renewcommand{\arraystretch}{1.12}
\scriptsize
\begin{tabular}{@{}p{0.27\textwidth}p{0.70\textwidth}@{}}
\toprule
Visual Input & GPT-5.6-Luna Description \\
\midrule
\textbf{PerceptionBench}\par\medskip
\centering\includegraphics[width=0.94\linewidth]{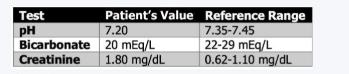}
&
The image is a rectangular table with a dark-gray header and three data rows. The first and third data rows have light-gray backgrounds, while the middle row is white. Its columns are ``Test,'' ``Patient's Value,'' and ``Reference Range.'' The rows report pH as 7.20 and 7.35--7.45; Bicarbonate as 20 mEq/L and 22--29 mEq/L; and Creatinine as 1.80 mg/dL and 0.62--1.10 mg/dL. \\
\midrule
\textbf{MMMU}\par\medskip
\centering\includegraphics[width=0.94\linewidth]{figures/oracle_description_examples/gpt_mmmu_75.png}
&
The image is a black-and-white chemical reaction diagram. ``Glyceraldehyde 3-phosphate + H$_2$O'' appears to the left of a horizontal reaction arrow, while ``3-phosphoglycerate'' appears below the product side. Above the arrow, NAD$^+$ and NADH + H$^+$ are connected by a curved arrow that bends downward and rises toward NADH + H$^+$. All text and lines are dark against a plain white background. \\
\midrule
\textbf{MathVista}\par\medskip
\centering\includegraphics[width=0.75\linewidth]{figures/oracle_description_examples/gpt_mathvista_88.png}
&
The image shows a large upright triangle centered on a white background. Its interior is uniformly light blue, with a brighter blue outline. The triangle is approximately equilateral, with a sharp top vertex and a horizontal base. Each of its three sides is labeled ``3 cm'' in dark-gray text, and no other objects or chart elements are visible. \\
\midrule
\textbf{MME}\par\medskip
\centering\includegraphics[width=0.75\linewidth]{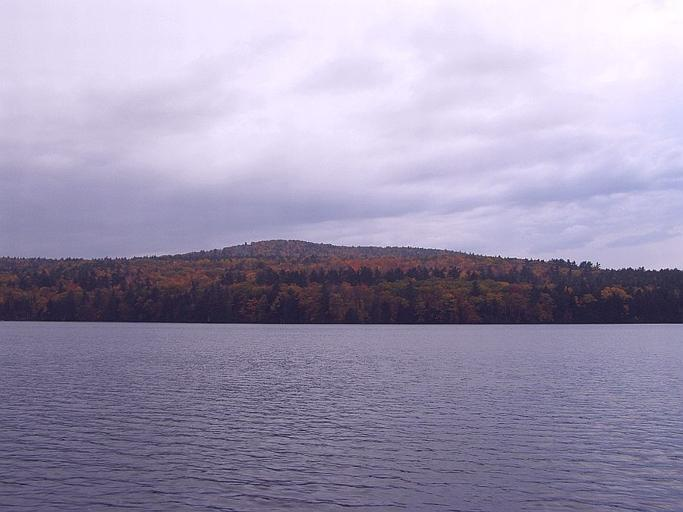}
&
The image is a wide photograph of a lake under an overcast sky. Layered gray and lavender clouds fill the upper half. A tree-covered shoreline and low rounded ridge extend across the center, with autumn foliage in green, brown, rust, orange, and muted red. Blue-gray water with fine horizontal ripples fills the foreground; no people, buildings, or boats are visible. \\
\bottomrule
\end{tabular}
\end{table*}

\begin{table*}[h]
\centering
\caption{\textbf{Example of perception recovery.} Qwen3-VL-4B answers the same MMMU question before and after receiving the question-blind description generated by GPT-5.6-Luna, recovering the perception failure.}
\label{tab:perception-recovery-example}
\setlength{\tabcolsep}{5pt}
\renewcommand{\arraystretch}{1.10}
\scriptsize
\begin{tabular}{@{}p{0.98\textwidth}@{}}
\toprule
\textbf{Visual Input} \\
\addlinespace[2pt]
{\centering
\includegraphics[width=0.31\linewidth]{figures/self_description_examples/mmmu_failure_39.png}
\par\smallskip
What is the size of real GDP in the short-run equilibrium?
\par\smallskip
(A) 500\qquad (B) 450\qquad (C) 400\qquad (D) 600
\par} \\
\midrule
\textbf{Base Model Generation} \\
\addlinespace[2pt]
The short-run equilibrium occurs where the aggregate demand (AD) curve intersects the short-run aggregate supply (SRAS) curve. The intersection appears at $(500,600)$, so the size of real GDP is 500 billion. Therefore, the answer is $\boxed{\mathrm{A}}$. \\
\midrule
\textbf{GPT-5.6-Luna Description} \\
\addlinespace[2pt]
The image is a square economic graph on a white background. The vertical axis is labeled ``Price Level,'' and the horizontal axis is labeled ``Real GDP.'' Both axes have tick marks and numeric labels at intervals of 100. A blue downward-sloping line labeled ``AD'' begins at approximately $(0,1000)$, passes through approximately $(400,600)$, and crosses the horizontal axis at approximately $(1000,0)$. A red upward-sloping line labeled ``SRAS'' begins at approximately $(0,200)$ and passes through approximately $(400,600)$. A vertical black line labeled ``LRAS'' is located at real GDP 500. \\
\midrule
\textbf{Recovered Answer} \\
\addlinespace[2pt]
The short-run equilibrium occurs where the AD and SRAS curves intersect. The description places their intersection at $(400,600)$, so the size of real GDP is 400 billion. Therefore, the answer is $\boxed{\mathrm{C}}$. \\
\bottomrule
\end{tabular}
\end{table*}

For the perception intervention, each oracle first generates a question-blind description. The target model then answers from the image augmented with the same description. A previous failure is included in $P_i^{M,B}$ if the base model succeeds with either oracle model description. For the reasoning intervention, the target model first generates its own question-blind description, with which the oracle answers the original question.
A previous failure is included in $R_i^{M,B}$ only if the oracle also answers correctly from the original image. We take the item-level union over Claude Opus 4.8 and GPT-5.6-Luna as defined in Eq.~\ref{eq:para-recovery}.

Descriptions are generated with a $1024$ token budget. Target model answers use up to $4096$ tokens. Decoding is greedy for the locally served models. Claude Opus 4.8 and GPT-5.6-Luna use the default configurations, with no explicit stop strings used.

\subsection{Reasoning Share}
Beyond the directional $\mathrm{BI}_{P}$ and $\mathrm{BI}_{R}$ reported in the main paper, we further report Reasoning Share (RS) in Tab.~\ref{tab:para-rs} as an alternative normalized comparison of the two measured headrooms.
\begin{equation}
\mathrm{RS}^{M,B}=\frac{|R^{M,B}|}{|P^{M,B}|+|R^{M,B}|}.
\label{eq:para-rs}
\end{equation}
RS directly measures their relative balance. $\mathrm{RS}=0.5$ when $P=R$, while values above $0.5$ indicate greater reasoning than perceptual headroom. This symmetric normalization is easy to interpret, but represents the two channels as complementary shares and therefore obscures their overlap. In contrast, the bottleneck indices $\mathrm{BI}_{P}$ and $\mathrm{BI}_{R}$ normalize each channel by $|P\cup R|$, measuring how broadly each intervention covers the distinct recoverable error set. Both indices can be high when the two interventions recover many of the same failures, thereby preserving information about their overlap. We therefore use $\mathrm{BI}$ in the main paper, while $\mathrm{RS}$ provides an alternative normalization that preserves the ordering and exhibits the same broad trends. The values below use the same results over the two oracles as Tab.~\ref{tab:para-main}.

\begin{table}[h]
\centering
\caption{\textbf{Reasoning Share across models.} RS reports $R/(P+R)$ for the four models and $17$
benchmarks in the main \score evaluation. $^{\dagger}$ represents multiple-choice benchmarks.}
\label{tab:para-rs}
\resizebox{\linewidth}{!}{%
\begin{tabular}{lrrrr}
\toprule
Benchmark & Qwen3-VL-2B & Qwen3-VL-4B & Llama-4-Scout-17B & Gemma-3-27B \\
\midrule
BLINK$^{\dagger}$ & 0.39 & 0.30 & 0.30 & 0.39 \\
VisNumBench$^{\dagger}$ & 0.43 & 0.44 & 0.46 & 0.42 \\
RealWorldQA & 0.37 & 0.50 & 0.40 & 0.38 \\
PerceptionBench & 0.31 & 0.24 & 0.33 & 0.32 \\
MMStar$^{\dagger}$ & 0.44 & 0.39 & 0.40 & 0.39 \\
ChartQA & 0.52 & 0.50 & 0.36 & 0.43 \\
HallusionBench & 0.43 & 0.40 & 0.34 & 0.36 \\
MME & 0.56 & 0.55 & 0.49 & 0.39 \\
MMBench$^{\dagger}$ & 0.50 & 0.49 & 0.43 & 0.45 \\
ScienceQA$^{\dagger}$ & 0.50 & 0.50 & 0.40 & 0.54 \\
MM-Vet & 0.41 & 0.37 & 0.41 & 0.46 \\
MMMU$^{\dagger}$ & 0.61 & 0.55 & 0.51 & 0.53 \\
MMMU-Pro$^{\dagger}$ & 0.49 & 0.48 & 0.56 & 0.60 \\
LogicVista$^{\dagger}$ & 0.50 & 0.44 & 0.42 & 0.28 \\
MathVista$^{\dagger}$ & 0.28 & 0.29 & 0.26 & 0.33 \\
MathVerse & 0.61 & 0.59 & 0.53 & 0.55 \\
MATH-Vision & 0.66 & 0.48 & 0.60 & 0.45 \\
\midrule
\textit{Mean} & 0.47 & 0.44 & 0.42 & 0.43 \\
\bottomrule
\end{tabular}}
\end{table}

\section{Experimental Details}
\label{app:experiments}

\subsection{Datasets and Implementation Details}

\paragraph{Training dataset.}
Following Vision-SR1~\citep{li2025visionsr1}, we train all methods on Vision-SR1-47K, which contains approximately $47$K image question--answer pairs collected from $24$ open-source datasets. The data span multimodal mathematical reasoning, commonsense knowledge, and general visual understanding. We use the same training split across all methods to isolate the effect of their learning objectives.

\paragraph{Benchmarks.}
We list all benchmarks evaluated in this work. The first seven benchmarks below are used in our post-training experiments, and we extend extra ten benchmarks for evaluation of \score.
\begin{itemize}
    \item \textbf{MME}~\citep{fu2023mme} evaluates both perceptual and cognitive abilities through a broad collection of visual tasks.
    \item \textbf{MMMU}~\citep{yue2024mmmu} assesses expert-level multimodal understanding and reasoning across academic disciplines.
    \item \textbf{MMMU-Pro}~\citep{yue2025mmmupro} strengthens MMMU with more challenging questions and redesigned multiple-choice settings.
    \item \textbf{MATH-Vision}~\citep{wang2024mathvision} evaluates mathematical problem solving grounded in diagrams and other visual contexts.
    \item \textbf{PerceptionBench}~\citep{lin2026perceptionbench} targets fine-grained atomic visual perception across diverse visual attributes.
    \item \textbf{ScienceQA}~\citep{lu2022scienceqa} evaluates multimodal science question answering using image and textual context.
    \item \textbf{VisNumBench}~\citep{weng2025visnumbench} tests numerical understanding and reasoning grounded in visual content.
    \item \textbf{BLINK}~\citep{fu2024blink} evaluates core visual perception skills, including depth, correspondence, and multi-view reasoning.
    \item \textbf{ChartQA}~\citep{masry2022chartqa} measures visual and logical reasoning over real-world charts.
    \item \textbf{HallusionBench}~\citep{guan2024hallusionbench} diagnoses visual illusion, language hallucination, and image-context reasoning in VLMs.
    \item \textbf{LogicVista}~\citep{xiao2024logicvista} evaluates visually grounded logical reasoning across multiple reasoning types.
    \item \textbf{MathVerse}~\citep{zhang2024mathverse} tests whether multimodal models understand and use diagrams when solving visual mathematics problems.
    \item \textbf{MathVista}~\citep{lu2024mathvista} evaluates mathematical reasoning across charts, diagrams, figures, and other visual contexts.
    \item \textbf{MMBench}~\citep{liu2024mmbench} provides a systematic multiple-choice evaluation of broad multimodal capabilities.
    \item \textbf{MMStar}~\citep{chen2024mmstar} evaluates six core multimodal capabilities using vision-dependent samples with reduced data leakage.
    \item \textbf{MM-Vet}~\citep{yu2023mmvet} evaluates integrated capabilities that require combinations of recognition, OCR, knowledge, generation, mathematics, and spatial understanding.
    \item \textbf{RealWorldQA}~\citep{xai2024realworldqa} measures spatial understanding of real world scenes.
\end{itemize}
We use each benchmark's official metric and evaluation protocol. MME is reported on its original scale in the result tables and normalized to a $100$-point scale only when computing the cross-benchmark average.

\paragraph{Training setup.}
We perform full-parameter fine-tuning of Qwen3-VL-2B-Instruct and Qwen3-VL-4B-Instruct on four NVIDIA H200 GPUs using fully sharded data parallelism and gradient checkpointing. We train with one PPO epoch per batch, a global batch size of $128$, and eight rollouts per prompt. We report the results of the 270 step checkpoints.
We use AdamW with a learning rate of $1\times 10^{-6}$, weight decay $0.01$, and maximum gradient norm $1.0$. 
Following GRPO, we use a standard binary accuracy reward from a rule-based verifier, rather than
self-rewarding signals based on model judgments or reasoning consistency
~\citep{yuan2024selfrewarding,zhang2025covo}. The final answer is extracted from the generated
\verb|\boxed{}| and compared with the ground-truth answer.
Following the official Vision-SR1 implementation, we use a low-variance KL penalty with coefficient $\beta=0.01$, asymmetric clipping ratios $\epsilon_{\mathrm{low}}=0.2$ and $\epsilon_{\mathrm{high}}=0.3$. During training, rollouts are sampled with temperature $1.0$ and top-$p=0.99$, with a maximum response length of $4096$ tokens. We use a constant learning-rate schedule without warmup.
The attention loss coefficient $\lambda$ is $1.0$, and supervision is applied to the last decoder layer. All experiments use random seed $2026$, and input images are capped at $1{,}003{,}520$ pixels ($1280\times28\times28$).

\paragraph{Implementation details and efficiency.}
The student attention is recomputed with gradients during the current policy forward pass, requiring no additional forward pass. 
We use generated response tokens as queries and visual tokens as keys. Attention is first normalized over visual keys independently for each head and averaged uniformly across heads. Padding tokens are excluded and different response lengths are handled by response-wise pooling. Siblings share the same image and therefore the same visual token layout.

\subsection{Additional Ablation Results}
\label{app:experiment-ablation}

In Sec.~\ref{sec:ablation}, we discuss the effectiveness of our method components and the robustness of our default coefficient and supervision scope. In this section, we further provide more details.

\paragraph{Grounding penalty.}
The grounding variant in Tab.~\ref{tab:component-ablation} derives a visual support set $G_g$ from the best sibling's pooled attention distribution $\widehat p_g$, instead of the full attention map~\citep{chen2026mnsp}. Specifically, we sort the visual tokens by $\widehat p_g$ and select the smallest prefix whose cumulative probability above a threshold $\gamma$.
We then penalize the current rollout attention mass outside this support:
\begin{equation}
\mathcal L_{\mathrm{ground}}
=\frac{1}{|\mathcal I|}\sum_{(g,i)\in\mathcal I}w_i
\sum_{j\notin G_g}\bar p_i^{\theta}(j).
\label{eq:grounded-mass}
\end{equation}
The variant adds $\beta\mathcal L_{\mathrm{ground}}$ to the sibling alignment objective, using the same configurations as \model.  $G_g$ is a sample-specific subset of the existing visual tokens.
We report the coefficient comparison results in Tab.~\ref{tab:coefficient-sensitivity-2b}, and layer supervision comparison in Tab.~\ref{tab:attention-layers-2b}.

\begin{table}[h]
\centering
\scriptsize
\caption{\textbf{Sensitivity to the attention-loss coefficient.} We vary $\lambda$ on Qwen3-VL-2B.}
\label{tab:coefficient-sensitivity-2b}
\resizebox{\linewidth}{!}{%
\begin{tabular}{crrrrrrrr}
\toprule
$\lambda$ & MME & MMMU & MMMU-Pro & MathVis & PercepBench & ScienceQA & VisNum. & Avg. \\
\midrule
$0.2$ & \underline{2248.4} & \textbf{52.7} & 34.2 & 19.1 & 20.6 & \underline{90.1} & \underline{38.2} & \underline{47.9} \\
$1.0$ & \textbf{2290.4} & \underline{52.1} & \textbf{35.1} & \textbf{21.1} & \textbf{24.4} & \textbf{90.3} & \textbf{40.8} & \textbf{49.4} \\
$5.0$ & 2228.8 & 49.4 & \underline{34.5} & \underline{19.7} & \underline{20.9} & 88.8 & 36.1 & 47.0 \\
\bottomrule
\end{tabular}}
\end{table}

\begin{table}[h]
\centering
\scriptsize
\caption{\textbf{Effect of supervised attention layers.} We extend the final-layer objective to the top 4, 8, 16, or all layers of Qwen3-VL-2B.}
\label{tab:attention-layers-2b}
\resizebox{\linewidth}{!}{%
\begin{tabular}{lrrrrrrrr}
\toprule
 & MME & MMMU & MMMU-Pro & MathVis & PercepBench & ScienceQA & VisNum. & Avg. \\
\midrule
Final layer & \textbf{2290.4} & \textbf{52.1} & \textbf{35.1} & \textbf{21.1} & \textbf{24.4} & \textbf{90.3} & \textbf{40.8} & \textbf{49.4} \\
Top 4  & \underline{2259.6} & 48.0 & \underline{34.3} & 18.4 & 19.5 & 88.5 & \underline{36.3} & 46.5 \\
Top 8  & 2242.8 & \underline{50.9} & 33.9 & \underline{20.4} & \underline{20.1} & \underline{89.7} & 36.1 & \underline{47.3} \\
Top 16 & 2245.6 & 47.1 & 29.9 & 14.8 & 18.8 & 85.9 & 33.2 & 44.3 \\
\bottomrule
\end{tabular}}
\end{table}

\end{document}